\documentclass[10pt,twocolumn,letterpaper]{article}

\usepackage{cvpr}
\usepackage{times}
\usepackage{epsfig}
\usepackage{graphicx}
\usepackage{amsmath}
\usepackage{amssymb}
\usepackage{subfigure}
\usepackage{float}

\makeatletter
\usepackage{xspace}
\def\@onedot{\ifx\@let@token.\else.\null\fi\xspace}
\DeclareRobustCommand\onedot{\futurelet\@let@token\@onedot}

\newcommand{\figref}[1]{Fig\onedot~\ref{#1}}

\def\eg{\emph{e.g}\onedot} 
\def\ie{\emph{i.e}\onedot}

\def\etal{\emph{et al}\onedot}

\makeatother

\usepackage[breaklinks=true,bookmarks=false]{hyperref}

\cvprfinalcopy % *** Uncomment this line for the final submission

\def\cvprPaperID{} % *** Enter the CVPR Paper ID here

\begin{document}
%%%%%%%%% TITLE
\title{Semantic Image Inpainting with Deep Generative Models}

\author{Raymond A. Yeh\thanks{Authors contributed equally.}, \ Chen Chen\footnotemark[1], \ Teck Yian Lim, \\ Alexander G. Schwing, Mark Hasegawa-Johnson, Minh N. Do \\
University of Illinois at Urbana-Champaign\\
{\tt\small \{yeh17, cchen156, tlim11, aschwing, jhasegaw, minhdo\}@illinois.edu}
% For a paper whose authors are all at the same institution,
% omit the following lines up until the closing ``}''.
% Additional authors and addresses can be added with ``\and'',
% just like the second author.
% To save space, use either the email address or home page, not both
}

\maketitle
%\thispagestyle{empty}

%%%%%%%%% ABSTRACT
\begin{abstract}
% !TEX root = ../inpainting_final.tex
Semantic image inpainting is a challenging task where large missing regions have to be filled based on the available visual data. Existing methods which extract information from only a single image generally produce unsatisfactory results due to the lack of high level context.  In this paper, we propose a novel method for  semantic image inpainting, which generates the missing content by conditioning on the available data. Given a trained generative model, we search for the closest encoding of the corrupted image in the latent image manifold using our context and prior losses. This encoding is then passed through the generative model to infer the missing content. In our method, inference is possible irrespective of how the missing content is structured, while the state-of-the-art learning based method requires  specific information about the holes in the training phase. Experiments on three datasets show that our method successfully predicts information in large missing regions and achieves pixel-level photorealism, significantly outperforming the state-of-the-art methods.

\end{abstract}

%%%%%%%%% INTRODUCTION
\section{Introduction}
% !TEX root = ../inpainting_final.tex

\begin{figure}[t]
	\begin{center}
		\includegraphics[scale=0.96]{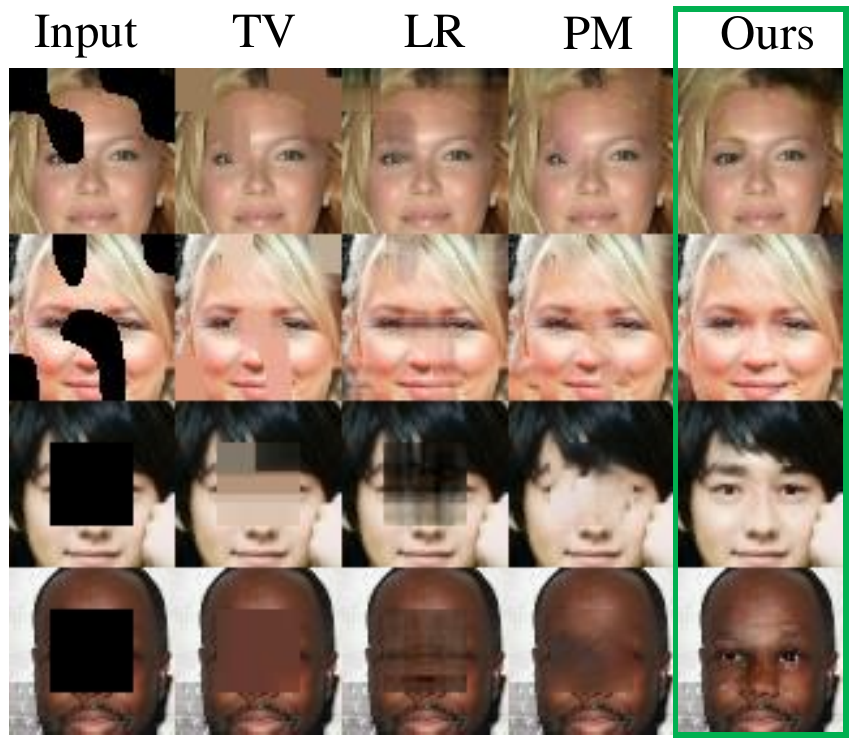}
		\caption{Semantic inpainting results by TV, LR, PM and our method. Holes are marked by black color.}
		\label{fig:traindition}
	\end{center}
	\vspace{-7mm}
\end{figure}

Semantic inpainting~\cite{pathak2016context} refers to the task of inferring arbitrary large missing regions in images based on image semantics. Since prediction of high-level context is required, this task is significantly more difficult than classical inpainting or image completion which is often more concerned with correcting spurious data corruption or removing entire objects. 
Numerous applications such as restoration of damaged paintings or image editing~\cite{bertalmio2000image} benefit from accurate semantic inpainting methods if large regions are missing. However, inpainting becomes increasingly more difficult if large regions are missing or if scenes are complex.

Classical inpainting methods are often based on either local or non-local information to
recover the image.  Most existing methods are designed for single image inpainting. Hence they are based on the  information available in the input image, and exploit  image priors to address the ill-posed-ness. 
For example, total variation (TV) based approaches~\cite{shen2002mathematical,afonso2011augmented} take into account the smoothness property of natural images, which is useful to fill small missing regions or remove spurious noise. Holes in textured images can be filled by finding a similar texture from the same image~\cite{efros1999texture}. Prior knowledge, such as statistics of patch offsets~\cite{he2012statistics}, planarity~\cite{huang2014image} or low rank (LR)~\cite{hu2013fast} can greatly improve the result as well. PatchMatch (PM)~\cite{barnes2009patchmatch} searches for similar patches in the available part of the image and quickly became one of the most successful inpainting methods due to its high quality and efficiency. However, all single image inpainting methods require appropriate information to be contained in the input image, \eg, similar pixels, structures, or patches. This assumption is hard to satisfy, if the missing region is large and possibly of arbitrary shape. Consequently, in this case, these methods are unable to recover the missing information. \figref{fig:traindition} shows some challenging examples with large missing regions, where local methods fail to recover the nose and eyes. 

In order to address inpainting in the case of large missing regions, non-local methods try to predict the missing pixels using external data. Hays and Efros~\cite{hays2007scene} proposed to cut and paste a semantically similar patch from a huge database. Internet based retrieval can be used to replace a target region of a scene~\cite{whyte2009get}. Both methods require exact matching from the database or Internet, and fail easily when the test scene is significantly different from any database image. Unlike previous hand-crafted matching and editing, learning based methods have shown promising results~\cite{mairal2008sparse,xie2012image,ren2015shepard,liu2016learning}. After an image dictionary or a neural network is learned, the training set is no longer required for inference. Oftentimes, these learning-based methods are designed for small holes or for removing small text in the image.

Instead of filling small holes in the image, we are interested in the more difficult task of semantic inpainting~\cite{pathak2016context}. It aims to predict the detailed content of a large region based on the context of surrounding pixels. A seminal approach for semantic inpainting, and closest to our work is the Context Encoder (CE) by Pathak \etal~\cite{pathak2016context}. Given a mask indicating missing regions, a neural network is trained to encode the context information and predict the unavailable content. However, the CE only takes advantage of the structure of holes during training but not during inference. Hence it results in blurry or unrealistic images especially when missing regions have arbitrary shapes.

In this paper, we propose a novel method for semantic image inpainting. We consider semantic inpainting as a constrained image generation problem and take advantage of the recent advances in generative modeling.
After a deep generative model, \ie, in our case an adversarial network~\cite{goodfellow2014generative,radford2015unsupervised}, is trained, we search for an encoding of the corrupted image that is ``closest'' to the image in the latent space. The encoding is then used to reconstruct the image using the generator. We define ``closest''  by a weighted context loss to condition on the corrupted image, and a prior loss to penalizes unrealistic images. Compared to the CE, one of the major advantages of our method is that it does not require the masks for training and 
can be applied for arbitrarily structured missing regions during inference.  We evaluate our method on three datasets: CelebA \cite{liu2015faceattributes}, SVHN \cite{netzer2011reading} and Stanford Cars \cite{krause20133d},  with different forms of missing regions. Results demonstrate that on challenging semantic inpainting tasks our method can obtain much more realistic images than the state of the art techniques.

%%%%%%%%% RELATED WORK
\section{Related Work}
%!TEX root = ../inpainting_camera_ready.tex

A large body of literature exists for image inpainting, and due to space limitations we are unable to discuss all of it in detail.
Seminal work in that direction includes the aforementioned works and references therein.
Since our method is based on generative models and deep neural nets, we will review the technically related learning based work in the following.

\noindent\textbf{Generative Adversarial Networks} (GANs) are a framework for training generative parametric models, and have been shown to produce high quality images~\cite{goodfellow2014generative, denton2015deep, radford2015unsupervised}. This framework trains two networks, a generator, $G$, and a discriminator $D$. $G$ maps a random vector $\mathbf{z}$, sampled from a prior distribution $p_{\mathbf{Z}}$, to the image space while $D$ maps an input image to a likelihood. The purpose of $G$ is to generate realistic images, while $D$ plays an adversarial role, discriminating between the image generated from $G$, and the real image sampled from the data distribution $p_{data}$. 

The $G$ and $D$ networks are trained by optimizing the loss function:
\begin{align} \label{gan_loss}
\min \limits_{G} \max \limits_{D} V(G,D) = & \mathbb{E}_{\mathbf{h} \sim p_{data}(\mathbf{\mathbf{h}})} [\log(D(\mathbf{h}))] +  \notag \\ &\mathbb{E}_{\mathbf{z} \sim p_\mathbf{Z}(\mathbf{z})} [\log(1-D(G(\mathbf{z}))],
\end{align}
where $\mathbf{h}$ is the sample from the $p_{data}$ distribution;  $\mathbf{z}$ is a random  encoding on the latent space.

With some user interaction, GANs have been applied in interactive image editing~\cite{zhu2016generative}. 
However,  GANs can not be directly applied to the inpainting task, because they produce an entirely unrelated image with high probability, unless constrained by the provided corrupted image. 

\begin{figure}[t]
	\begin{center}
		\includegraphics[scale=0.75]{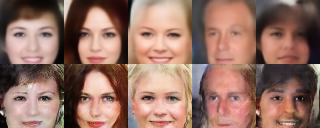}
		\caption{Images generated by a VAE and a DCGAN. First row: samples from a VAE. Second row: samples from a DCGAN.}
		\label{fig:vae}
	\end{center}
	\vspace{-7mm}
\end{figure}

\noindent\textbf{Autoencoders} and Variational Autoencoders (VAEs)~\cite{kingma2014auto} have become a popular approach to learning of complex distributions in an unsupervised setting. A variety of VAE flavors exist, \eg, extensions to attribute-based image editing tasks~\cite{yan2015attribute2image}. Compared to GANs, VAEs tend to generate overly smooth images, which is not preferred for inpainting tasks. \figref{fig:vae} shows some examples generated by a VAE and a Deep Convolutional GAN (DCGAN)~\cite{radford2015unsupervised}. Note that the DCGAN generates much sharper images. Jointly training VAEs with an adveserial loss prevents the smoothness~\cite{larsen2016autoencoding}, but may lead to artifacts. 

The Context Encoder (CE)~\cite{pathak2016context} can be also viewed as an autoencoder conditioned on the corrupted images.

It produces impressive reconstruction results when the structure of holes is fixed during both training and inference, \textit{e.g.}, fixed in the center,  but is less effective for arbitrarily structured regions. 

\begin{figure*}[htbp!]
	\begin{center}
		\includegraphics[scale=0.53]{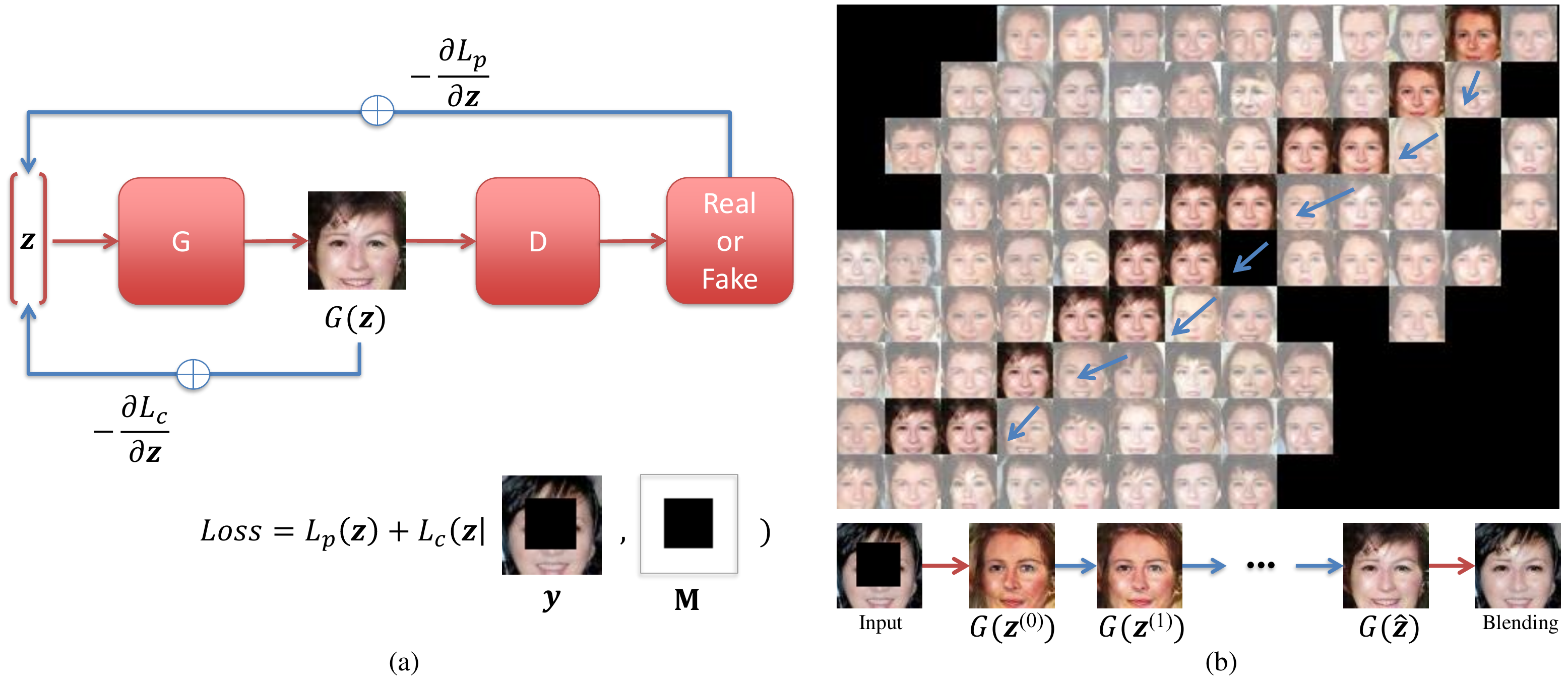} 
		\caption{The proposed framework for inpainting. (a) Given a GAN model trained on real images, we iteratively update $\mathbf{z}$ to find the closest mapping on the latent image manifold, based on the desinged loss functions. (b) Manifold traversing when iteratively updating $\mathbf{z}$ using back-propagation. $\mathbf{z}^{(0)}$ is random initialed; $\mathbf{z}^{(k)}$ denotes the result in $k$-th iteration; and $\hat{\mathbf{z}}$ denotes the final solution.}
		\label{fig:manifold}
	\end{center}
\end{figure*}

\noindent\textbf{Back-propagation to the input data} is employed in our approach to find the encoding which is close to the provided but corrupted image. In earlier work, back-propagation to augment data has been used for texture synthesis and style transfer~\cite{gatys2016image,gatys2015texture,li2016combining}. Google's DeepDream uses back-propagation to create dreamlike images~\cite{mordvintsev2015inceptionism}. 
Additionally, back-propagation has also been used to visualize and understand the learned features in a trained network, by ``inverting'' the network through updating the gradient at the input layer~\cite{mahendran2015understanding, dosovitskiy2015inverting, simonyan2013deep, linden1989inversion}. Similar to our method, all these back-propagation based methods require specifically designed loss functions for the particular tasks.

%%%%%%%%% METHOD
\section{Semantic Inpainting by Constrained Image Generation }
%!TEX root = ../inpainting_camera_ready.tex

To fill large missing regions in images, our method for image inpainting utilizes the generator $G$ and the discriminator $D$, both of which are trained with uncorrupted data. After training, the generator $G$ is able to take a point $\mathbf{z}$ drawn from $p_{\mathbf{Z}}$ and generate an image mimicking samples from $p_{data}$. We hypothesize that if $G$ is efficient in its representation then an image that is not from $p_{data}$ (\textit{e.g.}, corrupted data) should not lie on the learned encoding manifold, $\mathbf{z}$. Therefore, we aim to recover the encoding $\hat{\mathbf{z}}$ ``closest'' to the corrupted image while being constrained to the manifold, as illustrated in \figref{fig:manifold}; we visualize the latent manifold, using t-SNE \cite{maaten2008visualizing} on the 2-dimensional space, and the intermediate results in the optimization steps of finding $\hat{\mathbf{z}}$. After $\hat{\mathbf{z}}$ is obtained, we can generate the missing content by using the trained generative model $G$. 

More specifically, we  formulate the process of finding $\hat{\mathbf{z}}$ as an optimization problem. Let $\mathbf{y}$ be the corrupted image and $\mathbf{M}$ be the binary mask with size equal to the image, to indicate the missing parts. An example of $\mathbf{y}$ and $\mathbf{M}$ is shown in Fig. \ref{fig:manifold} (a).

Using this notation we define the ``closest'' encoding $\hat{\mathbf{z}}$ via:
\begin{equation}\label{eq:total}
\hat{\mathbf{z}} = \arg \min_{\mathbf{z}} \{ \mathcal{L}_{c}(\mathbf{z}| \mathbf{y}, \mathbf{M}) + \mathcal{L}_{p}(\mathbf{z}) \},
\end{equation}
where $\mathcal{L}_{c}$ denotes the context loss, which constrains the generated image given the input corrupted image $\mathbf{y}$ and the hole mask $\mathbf{M}$; $\mathcal{L}_{p}$ denotes the prior loss, which penalizes unrealistic images. The details of the proposed loss function will be discussed in the following sections.

Besides the proposed method, one may also consider using $D$ to update $\mathbf{y}$ by maximizing $D(\mathbf{y})$, similar to back-propagation in DeepDream~\cite{mordvintsev2015inceptionism} or neural style transfer~\cite{gatys2016image}. 
However, the corrupted data $\mathbf{y}$ is neither drawn from a real image distribution nor the generated image distribution.
Therefore, maximizing $D(\mathbf{y})$ may lead to a solution that is far away from the latent image manifold,
which may hence lead to results with poor quality.

\subsection{Importance Weighted Context Loss}
To fill large missing regions, our method takes advantage of the remaining available data. We designed the context loss to capture such information. A convenient choice for the context loss is simply the $\ell_2$ norm between the generated sample $G(\mathbf{z})$ and the uncorrupted portion of the input image $\mathbf{y}$. However, such a loss treats each pixel equally, which is not desired. 
Consider the case where the center block is missing: a large portion of the loss will be from pixel locations that are far away from the hole, such as the background behind the face. Therefore, in order to find the correct encoding, we should pay significantly more attention to the missing region that is close to the hole. 

To achieve this goal, we propose a context loss with the hypothesis that the importance of an uncorrupted pixel is positively correlated with the number of corrupted pixels surrounding it. A pixel that is very far away from any holes plays very little role in the inpainting process. We capture this intuition with the importance weighting term, $\mathbf{W}$,

\begin{equation} \label{eq:imporatnce_w}
\mathbf{W}_i = \left\{ \begin{matrix}
\sum\limits_{j \in N(i)} \frac{(1-\mathbf{M}_j)}{|N(i)|} & \text{if} \ \ \mathbf{M}_i \neq 0 \\ 
0 &  \text{if} \ \   \mathbf{M}_i = 0
\end{matrix}\right.,
\end{equation}
where $i$ is the pixel index, $\mathbf{W}_i$ denotes the importance weight at pixel location $i$, $N(i)$ refers to the set of neighbors of pixel $i$ in a local window, and $|N(i)|$ denotes the cardinality of $N(i)$. We use a window size of 7 in all experiments.

Empirically, we also found the $\ell_1$-norm to perform slightly better than the $\ell_2$-norm in our framework. Taking it all together, we define the conextual loss to be a weighted $\ell_1$-norm difference between the recovered image and the uncorrupted portion, defined as follows, 
\begin{equation} \label{eq:w_L_context}
\mathcal{L}_{c}(\mathbf{z}| \mathbf{y}, \mathbf{M}) =  \|\mathbf{W}\odot (G(\mathbf{z})-\mathbf{y})\|_1.
\end{equation}
Here, $\odot$ denotes the element-wise multiplication.

\subsection{Prior Loss}

The prior loss refers to a class of penalties based on high-level image feature representations instead of pixel-wise differences. In this work, the prior loss encourages the recovered image to be similar to the samples drawn from the training set. Our prior loss is different from the one defined in \cite{johnson2016perceptual} which uses features from pre-trained neural networks.

\begin{figure}[h]
	\centering
	\begin{tabular}{@{}cccc@{}}
		Real & Input & Ours w/o $\mathcal{L}_p$  & Ours w $\mathcal{L}_p$  \\
		\includegraphics[scale=0.8]{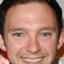}   &
		\includegraphics[scale=0.8]{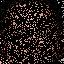} &
		\includegraphics[scale=0.8]{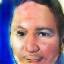} &
		\includegraphics[scale=0.8]{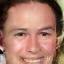}  \\
		\includegraphics[scale=0.8]{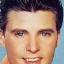}   &
		\includegraphics[scale=0.8]{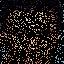} &
		\includegraphics[scale=0.8]{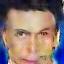} &
		\includegraphics[scale=0.8]{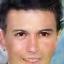}  \\	
	\end{tabular}
	\caption{Inpainting with and without the prior loss.}
	\label{fig:loss}
	\vspace{-3mm}
\end{figure}

Our prior loss penalizes unrealistic images. Recall that in GANs, the discriminator, $D$, is trained to differentiate  generated images from  real images. Therefore, we choose the prior loss to be identical to the GAN loss for training the discriminator $D$, \ie, 
\begin{equation}\label{eq:L_perceptual}
\mathcal{L}_{p}(\mathbf{z}) = \lambda \log(1-D(G(\mathbf{z}))).
\end{equation}
Here, $\lambda$ is a parameter to balance between the two losses. $\mathbf{z}$ is updated to fool $D$ and make the corresponding generated image more realistic. Without $\mathcal{L}_{p}$, the mapping from $\mathbf{y}$ to $\mathbf{z}$ may converge to a perceptually implausible result.
We illustrate this by showing the unstable examples where we optimized with and without $\mathcal{L}_{p}$ in Fig. \ref{fig:loss}. 

\subsection{Inpainting}
With the defined prior and context losses at hand, the corrupted image can be mapped to the closest $\mathbf{z}$ in the latent representation space, which we denote $\hat{\mathbf{z}}$. $\mathbf{z}$ is randomly initialized and updated using back-propagation on the total loss given in Eq. (\ref{eq:total}). Fig. \ref{fig:manifold} (b) shows for one example that $\mathbf{z}$ is approaching  the desired solution on the latent image manifold.

\begin{figure}[h]
	\centering
	\begin{tabular}{@{}cccc@{}}
		Overlay & Blend & Overlay  &  Blend  \\
		\includegraphics[scale=0.8]{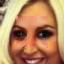}   &
		\includegraphics[scale=0.8]{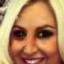} &
		\includegraphics[scale=0.8]{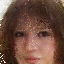} &
		\includegraphics[scale=0.8]{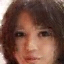}  
	\end{tabular}
	\caption{Inpainting with and without blending.}
	\label{fig:blend}
\end{figure}

After generating $G(\hat{\mathbf{z}})$, the inpainting result can be easily obtained by overlaying the uncorrupted pixels from the input.
However, we found that the predicted pixels may not exactly preserve the same intensities of the surrounding pixels, although the content is correct and well aligned. Poisson blending \cite{perez2003poisson} is used to reconstruct our final results. The key idea is to keep the gradients of $G(\hat{\mathbf{z}})$ to preserve image details while shifting the color to match the color in the input image $\mathbf{y}$. Our final solution, $\hat{\mathbf{x}}$, can be obtained by:
\begin{align}\label{eq:blend}
\hat{\mathbf{x}}&= \arg \min_\mathbf{x} \|\nabla \mathbf{x} - \nabla G(\hat{\mathbf{z}}) \|^2_2, \ \notag \\
& \text{s.t.} \ \mathbf{x}_i = \mathbf{y}_i \ \ \text{for} \ \ \mathbf{M}_i = 1,
\end{align}
where $\nabla$ is the gradient operator. The minimization problem contains a quadratic term, which has a unique solution \cite{perez2003poisson}. Fig. \ref{fig:blend} shows two examples where we can find visible seams without blending.

\subsection{Implementation Details}
In general, our contribution is orthogonal to specific GAN architectures and  our method can take advantage of any generative model $G$. We used the DCGAN model architecture from Radford \textit{et al.} \cite{radford2015unsupervised} in the experiments. The generative model, $G$, takes a random 100 dimensional vector drawn from a uniform distribution between $[-1,1]$ and generates a $64 \times 64 \times 3$ image.
The discriminator model, $D$, is structured essentially in reverse order. The input layer is an image of dimension $64\times 64 \times 3$, followed by a series of convolution layers where the image dimension is half, and the number of channels is double the size of the previous layer, and the output layer is a two class softmax.

For training the DCGAN model, we follow the training procedure in \cite{radford2015unsupervised} and use Adam \cite{kingma2014adam} for optimization. We choose $\lambda = 0.003$ in all our experiments. We also perform data augmentation of random horizontal flipping on the training images. In the inpainting stage, we need to find $\hat{\mathbf{z}}$ in the latent space using back-propagation. We use Adam for optimization and restrict $\mathbf{z}$ to $[-1,1]$ in each iteration, which we observe to produce more stable results. We terminate the back-propagation after 1500 iterations. We use the identical setting for all testing datasets and masks.

%%%%%%%%% EXPERIMENTS
\section{Experiments}
%!TEX root = ../inpainting_camera_ready.tex
In the following section we evaluate results qualitatively and quantitatively, more comparisons are provided in the supplementary material. 

\subsection{Datasets and Masks}

We evaluate our method on three dataset:  the CelebFaces Attributes Dataset (CelebA) \cite{liu2015faceattributes}, the Street View House Numbers (SVHN) \cite{netzer2011reading}  and the Stanford Cars Dataset \cite{krause20133d}.  

The CelebA contains $202,599$ face images with coarse alignment \cite{liu2015faceattributes}. We remove approximately 2000 images from the dataset for testing.  The images are cropped at the center to $64 \times 64$, which contain faces with various viewpoints and expressions. 

The SVHN dataset contains a total of 99,289 RGB images of cropped house numbers. The images are resized to $64 \times 64$ to fit the DCGAN model architecture. We used the provided training and testing split. The numbers in the images are not aligned and have different backgrounds. 

The Stanford Cars dataset contains 16,185 images of 196 classes of cars. Similar as the CelebA dataset, we do not use any attributes or labels for both training and testing. The cars are cropped based on the provided bounding boxes and resized to $64 \times 64$. As before, we use the provided training and test set partition. 

We test four different shapes of masks: 1) central block masks; 2) random pattern masks \cite{pathak2016context} in \figref{fig:traindition}, with approximately $25\%$ missing; 3) $80\%$ missing complete random masks; 4) half missing masks (randomly horizontal or vertical).

\subsection{Visual Comparisons}
\begin{figure}[t]
	\vspace{-0.5cm}
	\begin{center}
		\includegraphics[scale=0.95]{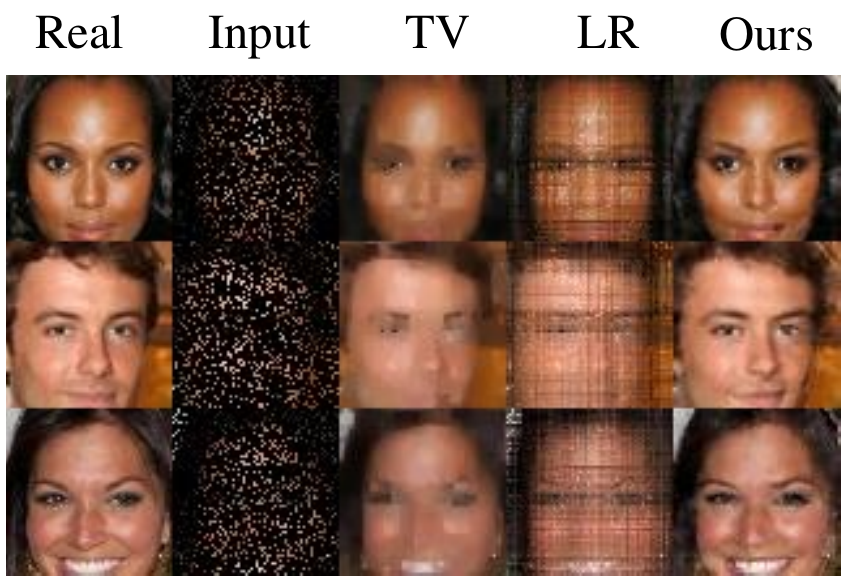}
		\caption{Comparisons with local inpainting methods TV and LR inpainting on examples with random $80\%$ missing. }
		\label{fig:comp_local}
	\end{center}
\end{figure}

\noindent\textbf{Comparisons with TV and LR inpainting.}
We compare our method with local inpainting methods. As we already showed in  \figref{fig:traindition}, local methods generally fail for large missing regions. We compare our method with TV inpainting \cite{afonso2011augmented} and LR inpainting \cite{lu2014generalized,hu2013fast} on images with small random holes. The test images and results are shown in Fig. \ref{fig:comp_local}.  Due to a large number of missing points, TV and LR based methods cannot recover enough image details, resulting in very blurry and noisy images. PM \cite{barnes2009patchmatch} cannot be applied to this case due to insufficient available patches.

\begin{figure}[t]
\vspace{-0.5cm}
	\begin{center}
		\includegraphics[scale=0.95]{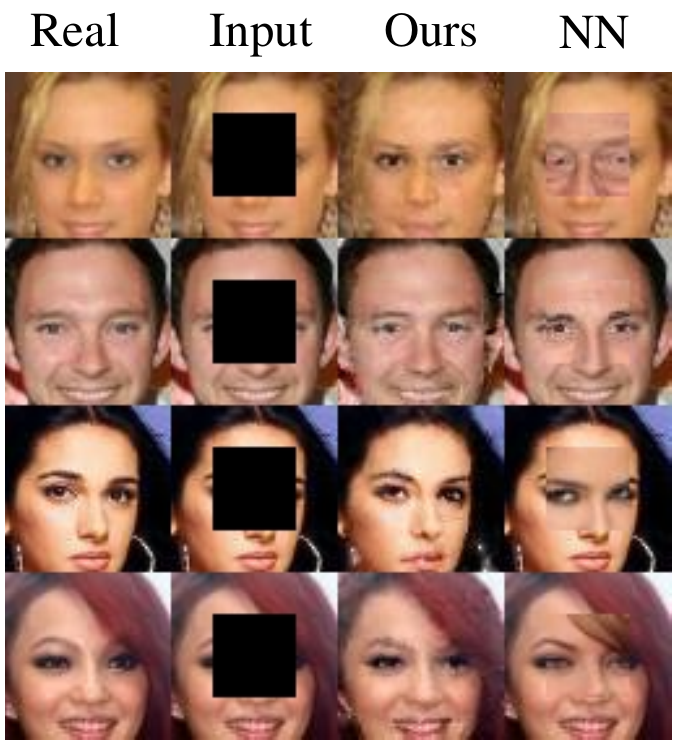}
		\caption{Comparisons with nearest patch retrieval. }
		\label{fig:nearest}
	\end{center}
	\vspace{-0.5cm}
\end{figure}

\noindent\textbf{Comparisons with NN inpainting.}
Next we compare our method with nearest neighbor (NN) filling from the training dataset, which is a key component in retrieval based methods \cite{hays2007scene,whyte2009get}. Examples are shown in Fig. \ref{fig:nearest}, where the misalignment of skin texture, eyebrows, eyes and hair can be clearly observed by using the nearest patches in Euclidean distance. Although people can use different features for retrieval, the inherit misalignment problem cannot be easily solved \cite{pathak2016context}.
Instead, our results are obtained automatically without any registration. 

\noindent\textbf{Comparisons with CE.}
In the remainder, we compare our result with those obtained from the CE \cite{pathak2016context}, the state-of-the-art method for semantic inpainting. It is important to note that the masks is required to train the CE. For a fair comparison, we use all the test masks in the training phase for the CE. However, there are infinite shapes and missing ratios for the inpainting task. To achieve satisfactory results one may need to re-train the CE.
In contrast, our method can be applied to arbitrary masks without re-training the network, which is according to our opinion a huge advantage when considering inpainting applications. 

\begin{figure}[t]
\vspace{-0.5cm}
	\centering
	\includegraphics[scale=1.09]{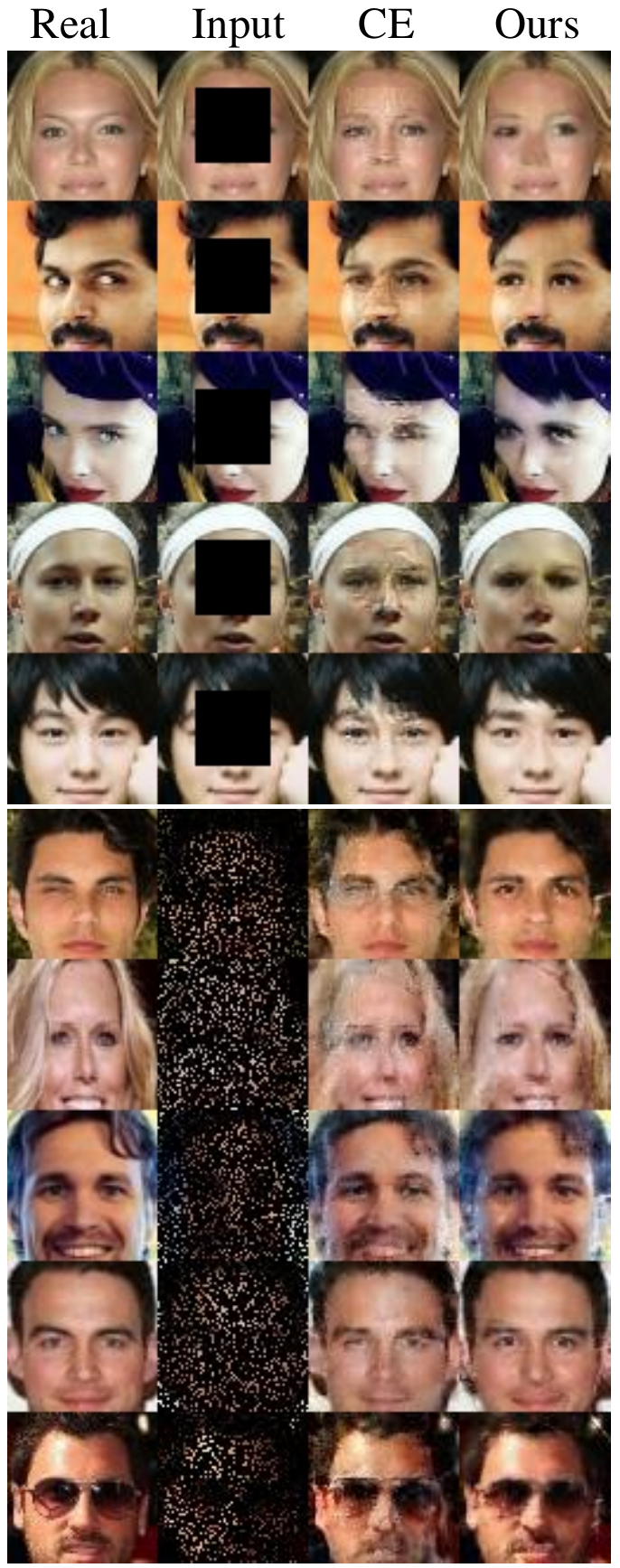}
	\vspace{-0.1cm}
	\caption{Comparisons with CE on the CelebA dataset. }
	\label{fig:celebA1}
	\vspace{-0.5cm}
\end{figure}

\begin{figure}[t]
\vspace{-0.5cm}
	\centering
	\includegraphics[scale=1.09]{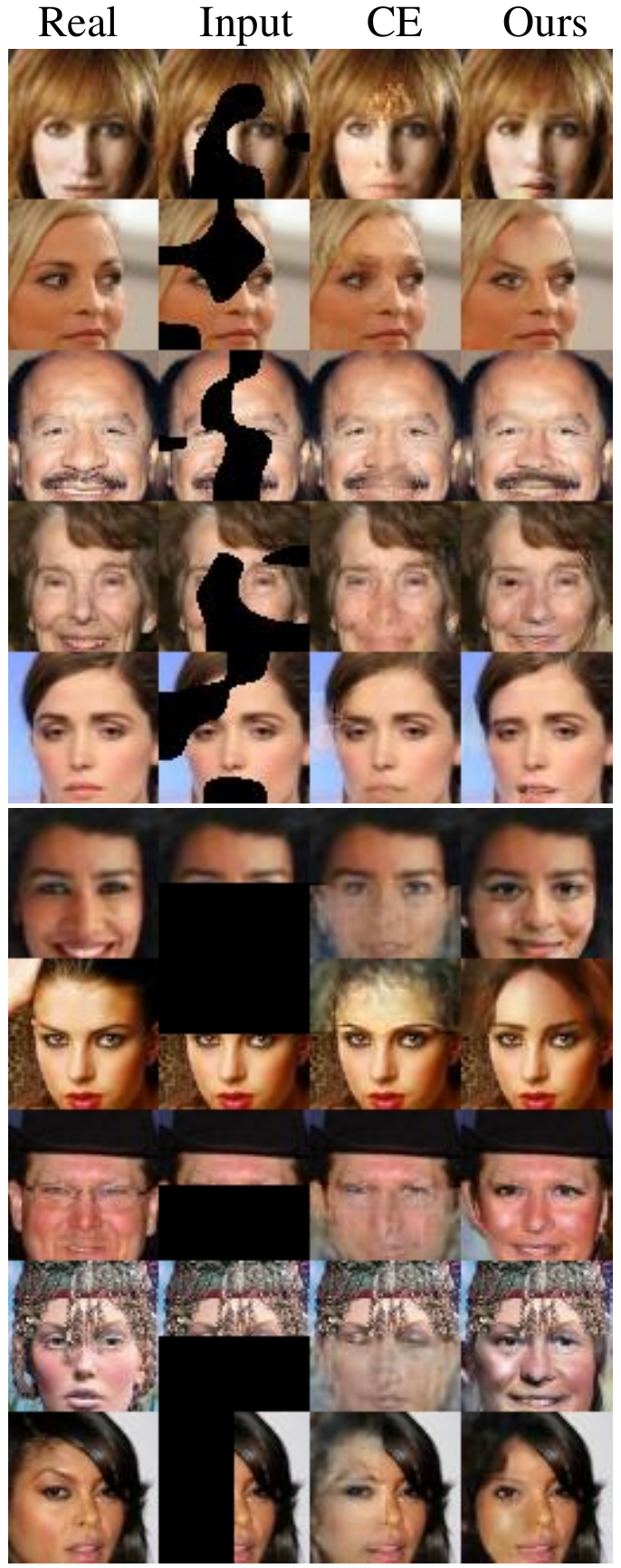}
	\caption{Comparisons with CE on the CelebA dataset. }
	\label{fig:celebA2}
	\vspace{-0.5cm}
\end{figure}

\begin{figure}[t]
\vspace{-0.5cm}
	\centering
		\includegraphics[scale=1.09]{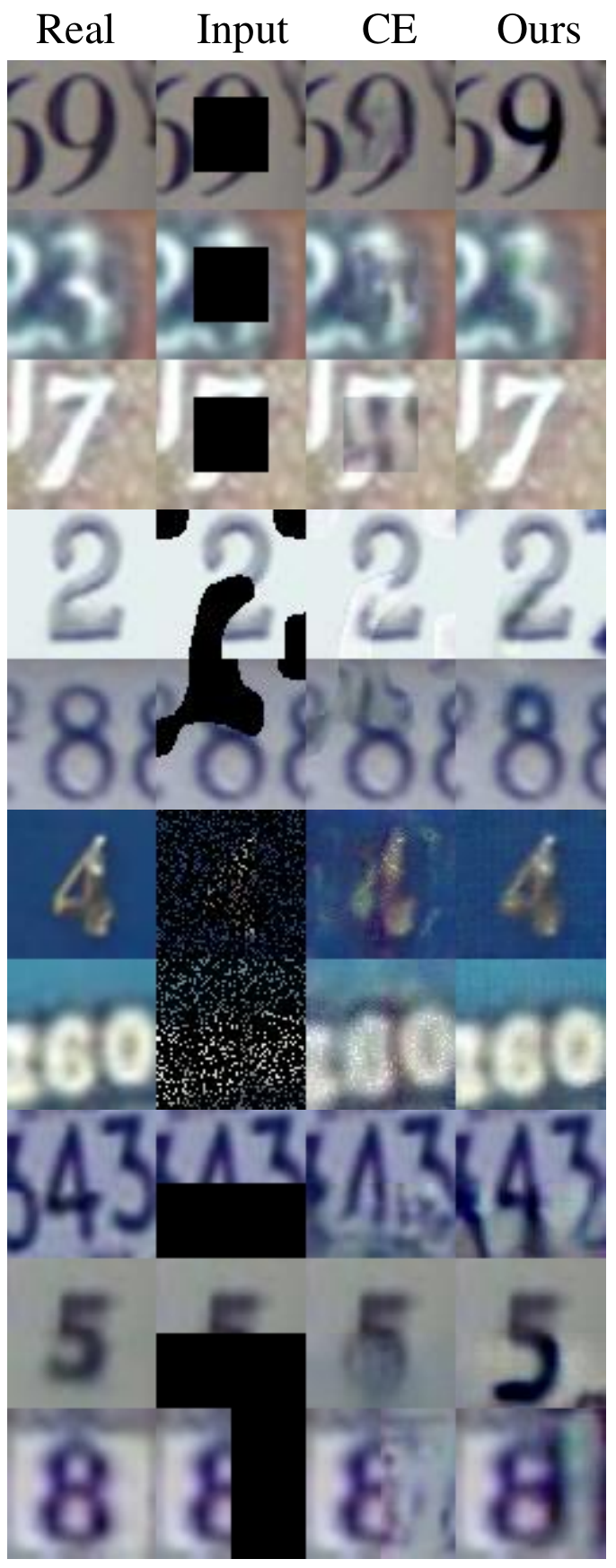}
	\caption{Comparisons with CE on the SVHN dataset. }
	\label{fig:SVHN}
	\vspace{-0.5cm}
\end{figure}

\begin{figure}[t]
\vspace{-0.5cm}
	\centering
		\includegraphics[scale=1.09]{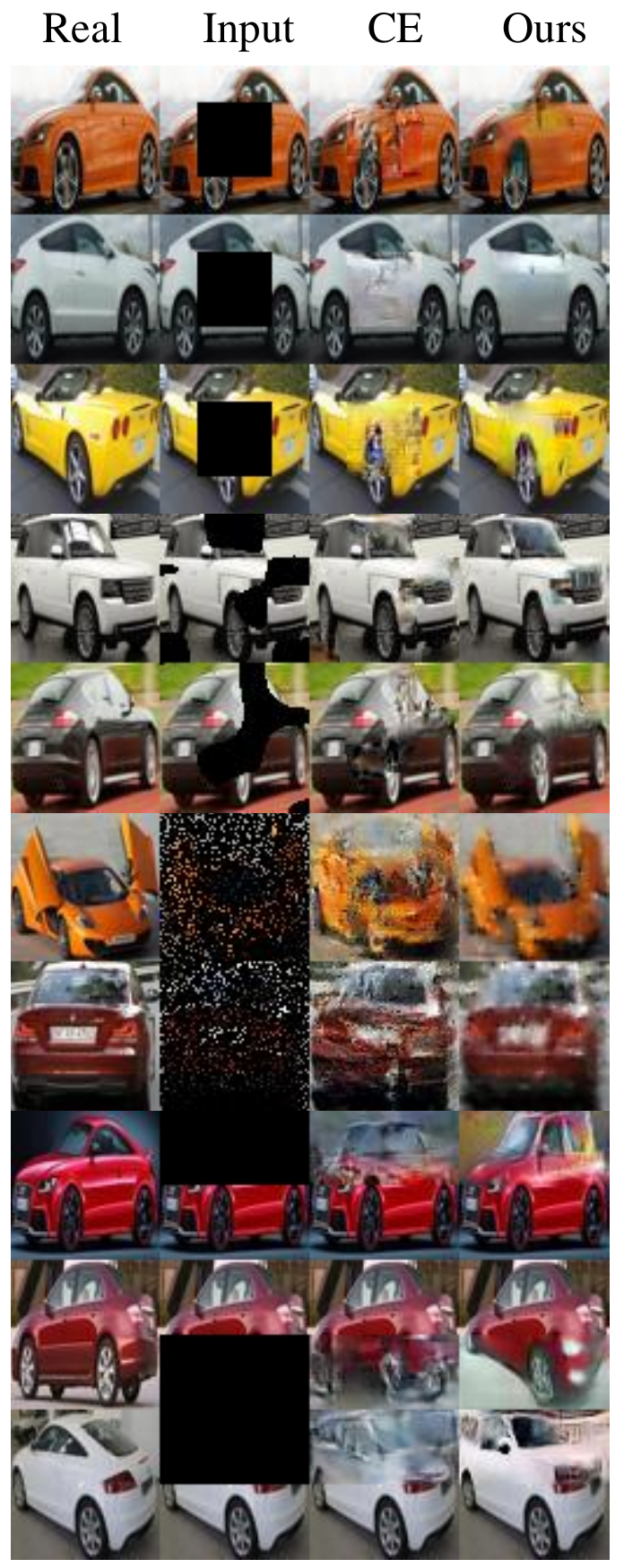} 
	\caption{Comparisons with CE on the car dataset.}
	\label{fig:car64}
	\vspace{-0.5cm}
\end{figure}

Figs.~\ref{fig:celebA1} and \ref{fig:celebA2} show the results on the CelebA dataset with four types of masks. Despite  some small artifacts,  the CE performs best with central masks. This is due to the fact that the hole is always fixed during both training and testing in this case, and the CE can easily learn to fill the hole from the context. However,  random missing data, is much more difficult for the CE to learn. 
In addition, the CE does not use the mask for inference but pre-fill the hole with the mean color. It may mistakenly treat some uncorrupted pixels with similar color as unknown.
We could observe that the CE has more artifacts and blurry results when the hole is at random positions. In many cases, our results are as realistic as the real images.  Results on SVHN and car datasets are shown in Figs.~\ref{fig:SVHN} and \ref{fig:car64}, and our method generally produces visually more appealing results than the CE since the images are sharper and contain fewer artifacts. 

\subsection{Quantitative Comparisons}

It is important to note that semantic inpainting is not trying to reconstruct the ground-truth image. The goal is to fill the hole with realistic content. Even the ground-truth image is one of many possibilities. However, readers may be interested in  quantitative results, often reported by classical inpainting approaches. 
Following previous work, we compare the PSNR values of our results and those by the CE. The real images from the dataset are used as groundtruth reference. Table~\ref{psnr} provides the results on the three datasets. The CE has higher PSNR values in most cases except for the random masks, as they are trained to minimize the mean square error. Similar results are obtained using SSIM \cite{wang2004image} instead of PSNR. These results  conflict with the aforementioned  visual comparisons, where our results generally yield to better perceptual quality.

\begin{table}[t]
	\caption{The PSNR values (dB) on the test sets. Left/right results are by CE\cite{pathak2016context}/ours.} \label{psnr}
	\centering
	\begin{tabular}{c|ccc}
		\hline
	Masks/Dataset	&  CelebA & SVHN & Cars   \\
		\hline
	Center   & \textbf{21.3}/19.4 & \textbf{22.3}/19.0 & \textbf{14.1}/13.5\\	
	pattern  &\textbf{ 19.2}/17.4 & \textbf{22.3}/19.8 & 14.0/\textbf{14.1}\\
	random   & 20.6/\textbf{22.8} & 24.1/\textbf{33.0} & 16.1/\textbf{18.9}\\
	half     & \textbf{15.5}/13.7 & \textbf{19.1}/14.6 & \textbf{12.6}/11.1\\
		\hline
\end{tabular}\end{table}

\begin{figure}[t]
		\vspace{-0.2cm}
		\centering
		\begin{tabular}{@{}ccc@{}}
			Input & CE & Ours \\
			\includegraphics[scale=1]{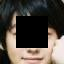}   &
			\includegraphics[scale=1]{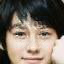} &
			\includegraphics[scale=1]{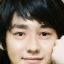}  \\
			Real & CE Error $\times$ 2	  & Ours Error $\times$ 2			 \\
			\includegraphics[scale=1]{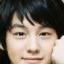} &
			\includegraphics[scale=1]{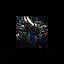} &
			\includegraphics[scale=1]{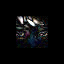} 
			
		\end{tabular}
		\caption{The error images for one example. The PSNR for context encoder and ours are 24.71 dB and 22.98 dB, respectively. The errors are amplified for display purpose. }
		\label{fig:errors}
	\end{figure}

We investigate this claim by carefully investigating the errors of the results. Fig.~\ref{fig:errors} shows the results of one example and the corresponding error images. Judging from the figure, our result looks artifact-free and very realistic, while the result obtained from the CE has visible artifacts in the reconstructed region. However, the PSNR value of CE is 1.73dB higher than ours. The error image shows that our result has large errors in hair area, because we generate a hairstyle which is different from the real image.  This indicates that quantitative result do not represent well the real performance of different methods when the ground-truth is not unique. Similar observations can be found in recent super-resolution works \cite{johnson2016perceptual,ledig2016photo}, where better visual results corresponds to lower PSNR values.

For random holes, both methods achieve much higher PSNR, even with $80\%$ missing pixels. In this case, our method outperforms the CE. This is because  uncorrupted pixels are spread across the entire image, and the flexibility of the reconstruction is strongly restricted; therefore PSNR is more meaningful in this setting which is more similar to the one considered in classical inpainting works.

\subsection{Discussion}

\begin{figure}[t]
	\vspace{-0.1cm}
	\begin{center}
		\includegraphics[scale=1]{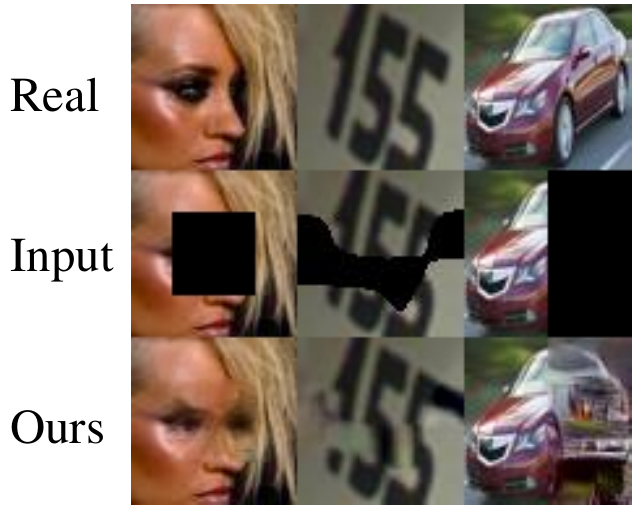}
		\caption{Some failure examples. }
		\label{fig:failure}
	\end{center}
	\vspace{-0.5cm}
\end{figure}

While the results are promising, the limitation of our method is also obvious. Indeed, its prediction performance strongly relies on the generative model and the training procedure. 
Some failure examples are shown in Fig.~\ref{fig:failure}, where our method cannot find the correct $\hat{\mathbf{z}}$ in the latent image manifold. The current GAN model in this paper works well for relatively simple structures like faces, but is too small to represent complex scenes in the world. Conveniently, stronger generative models,  improve our method in a straight-forward way.

%%%%%%%%% CONCLUSION
\section{Conclusion}
%!TEX root = ../inpainting_camera_ready.tex
In this paper, we  proposed a novel method for semantic inpainting. 
Compared to existing methods based on local image priors or patches, the proposed method learns the representation of training data, and can therefore predict meaningful content for corrupted images. Compared to CE, our method often obtains images with sharper edges which look much more realistic. Experimental results  demonstrated its superior performance on challenging image inpainting examples. 

\vspace{12pt}
\noindent\textbf{Acknowledgments:}
This work is supported in part by IBM-ILLINOIS Center for Cognitive Computing Systems Research (C3SR) - a research collaboration as part of the IBM Cognitive Horizons Network. This work is supported by NVIDIA Corporation with the donation of a GPU.

\clearpage

{\small
\bibliographystyle{ieee}
\bibliography{egbib}
}

\clearpage
%%%%%%%%% Supplementary
\section{Supplementary Material}
%!TEX root = ../inpainting_final.tex
\begin{figure}[h]
\centering
		\onecolumn\includegraphics[scale=1.1]{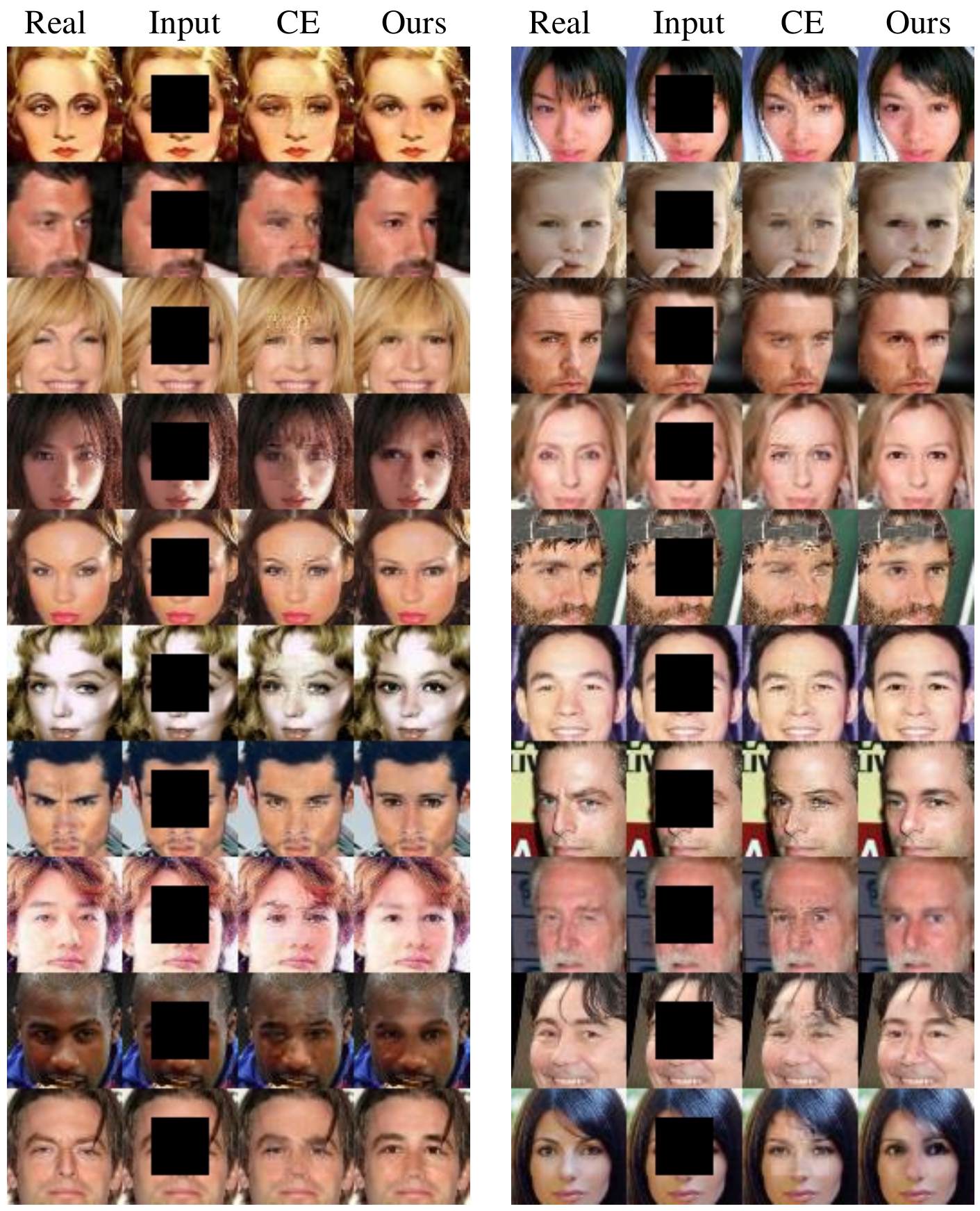}
		\caption{Additional results on the celebA dataset.}
\end{figure}

\begin{figure}[t]
\centering
		\onecolumn\includegraphics[scale=1.1]{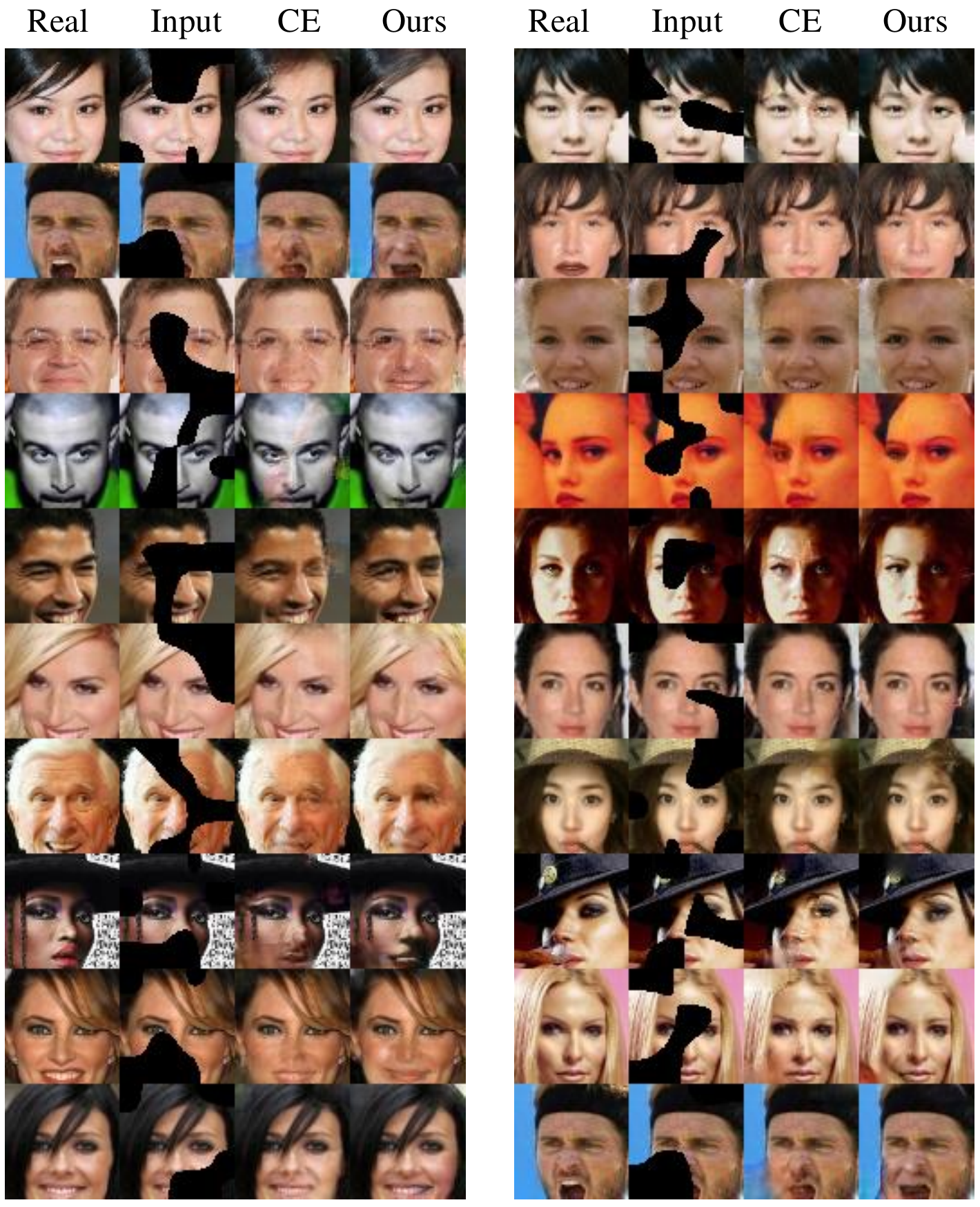}
		\caption{Additional results on the celebA dataset.}
\end{figure}

\begin{figure}[t]
\centering
		\onecolumn\includegraphics[scale=1.1]{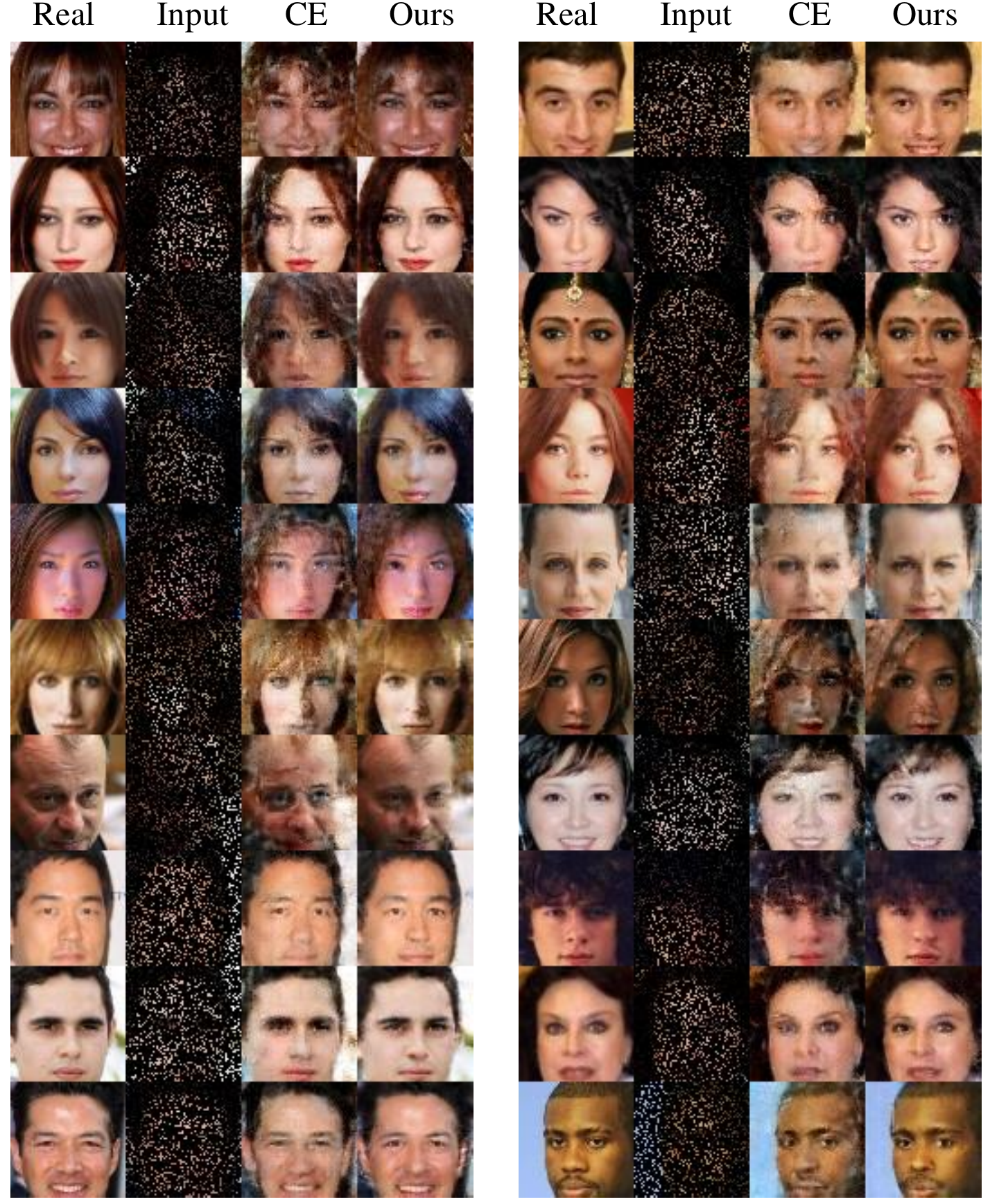}
		\caption{Additional results on the celebA dataset.}
\end{figure}

\begin{figure}[t]
\centering
		\onecolumn\includegraphics[scale=1.1]{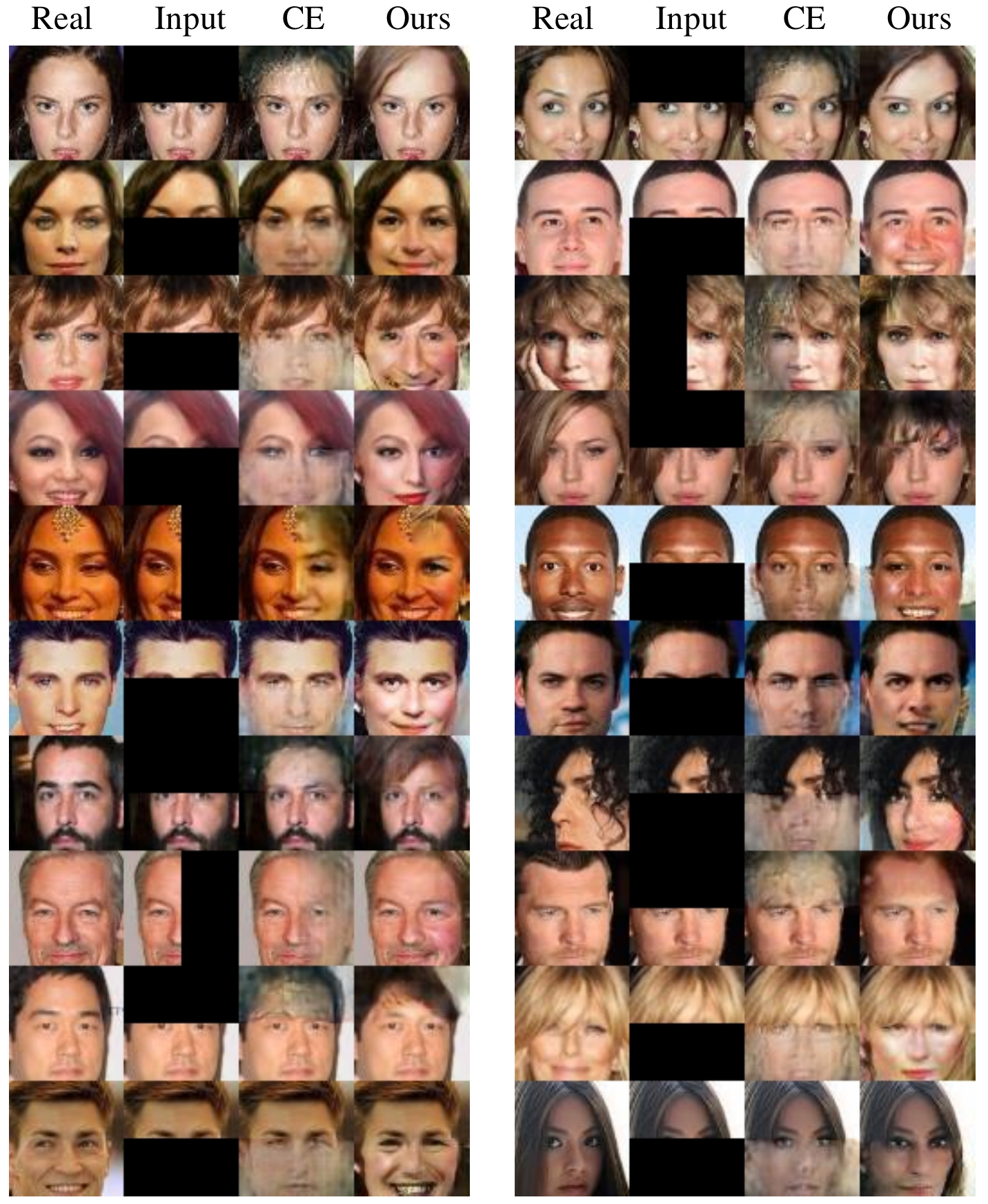}
		\caption{Additional results on the celebA dataset.}
\end{figure}

\begin{figure}[t]
\centering
		\onecolumn\includegraphics[scale=1.1]{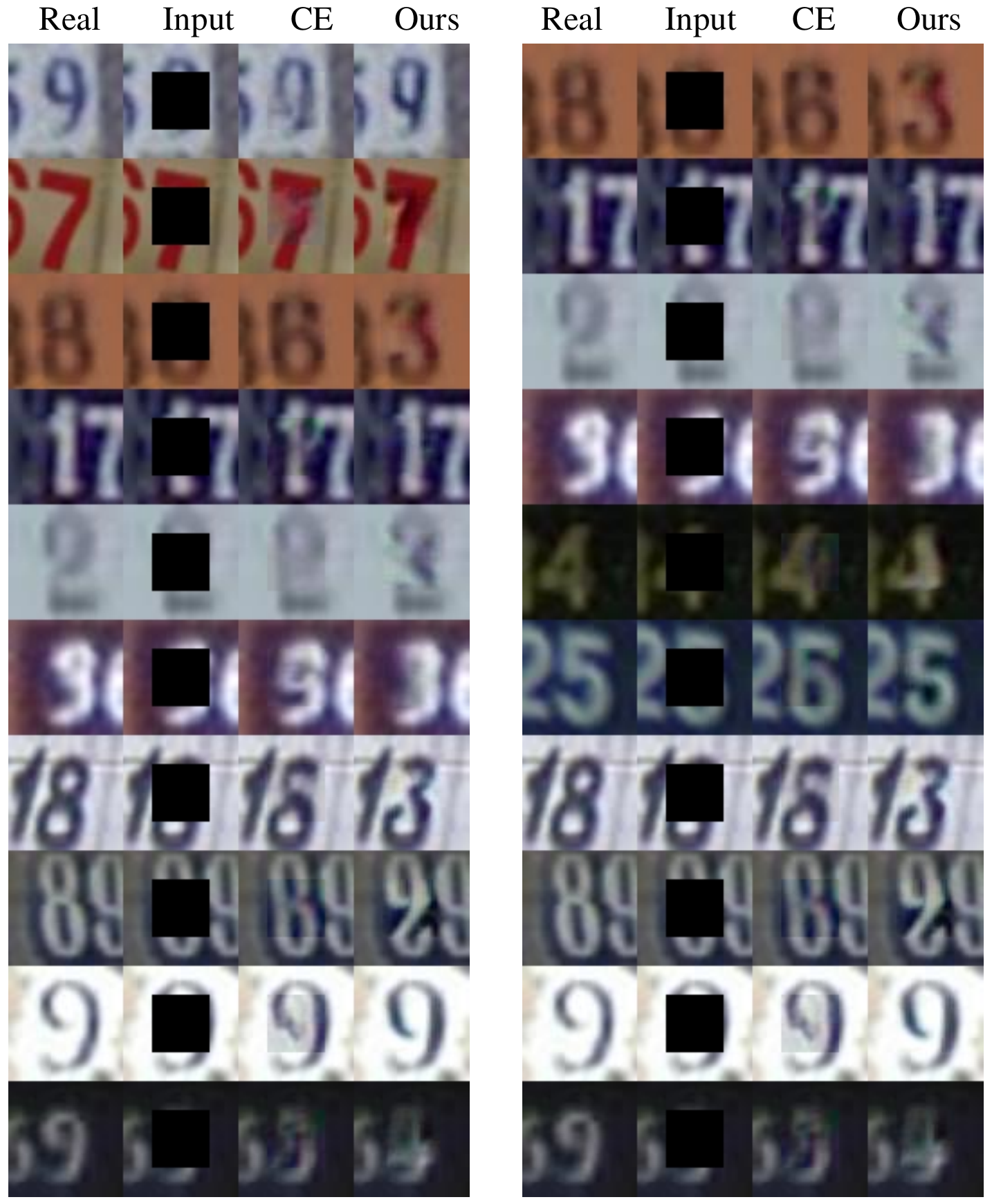}
		\caption{Additional results on the SVHN dataset.}
\end{figure}

\begin{figure}[t]
\centering
		\onecolumn\includegraphics[scale=1.1]{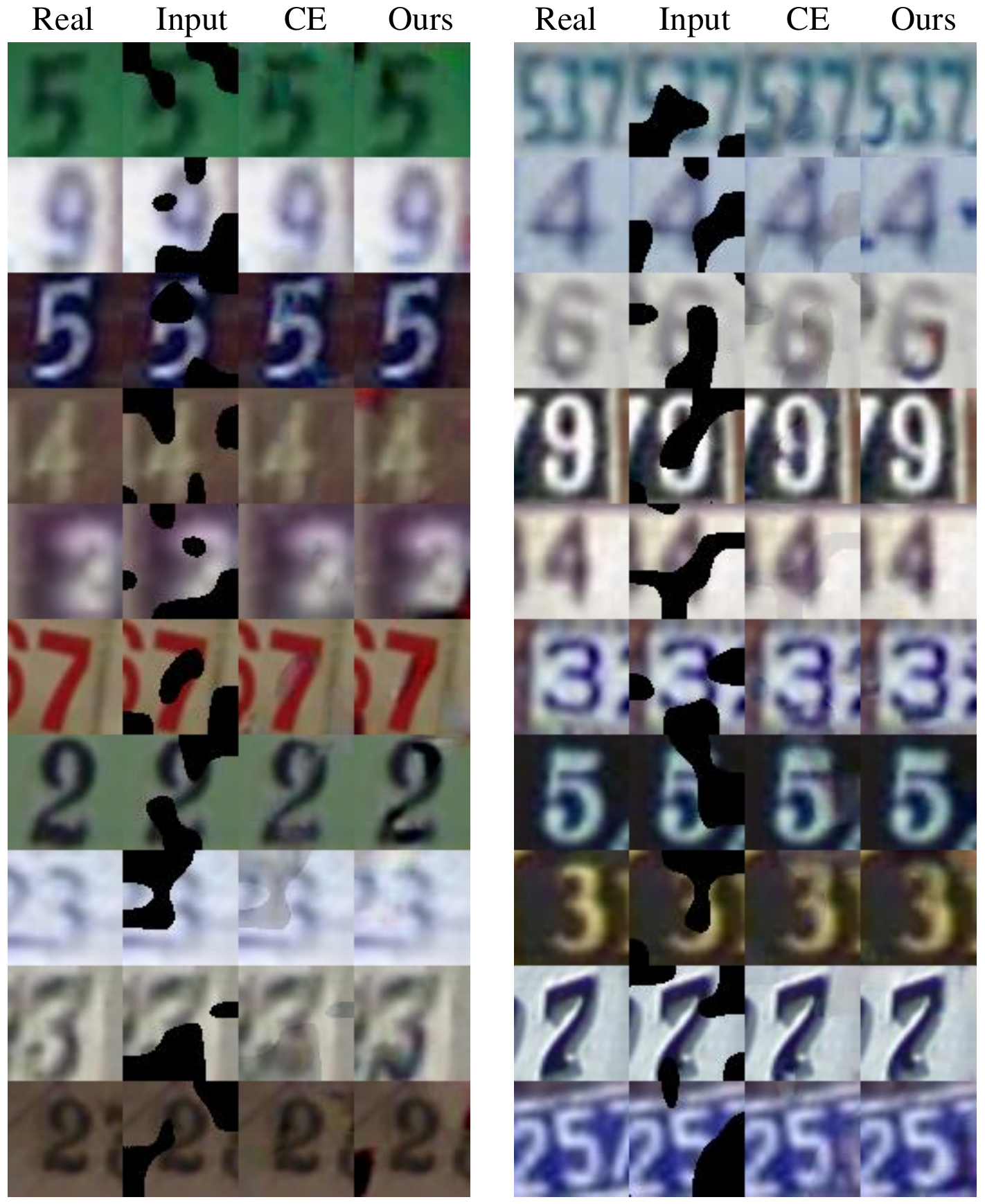}
		\caption{Additional results on the SVHN dataset.}
\end{figure}

\begin{figure}[t]
\centering
		\onecolumn\includegraphics[scale=1.1]{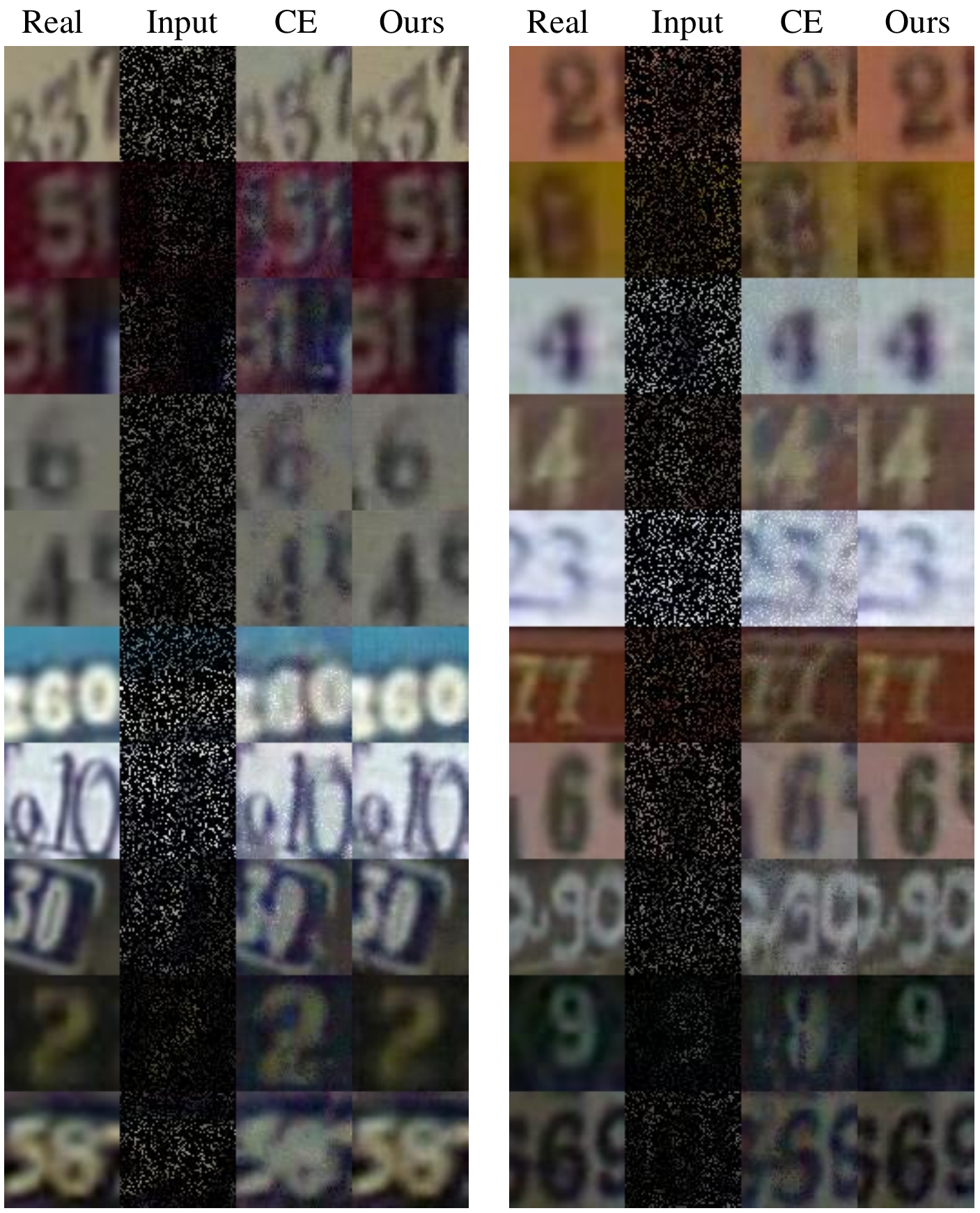}
		\caption{Additional results on the SVHN dataset.}
\end{figure}

\begin{figure}[t]
\centering
		\onecolumn\includegraphics[scale=1.1]{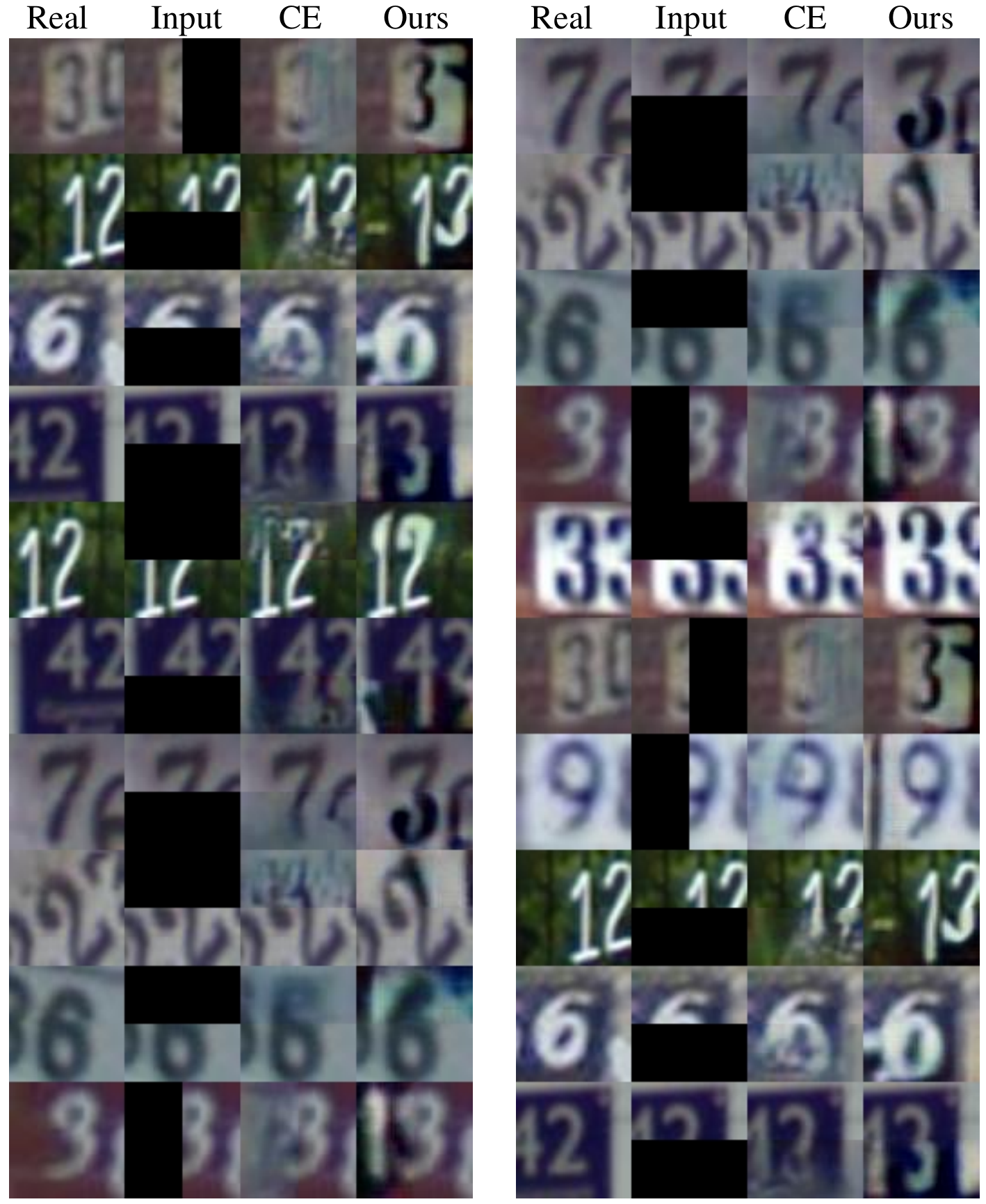}
		\caption{Additional results on the SVHN dataset.}
\end{figure}

\begin{figure}[t]
\centering
		\includegraphics[scale=1.1]{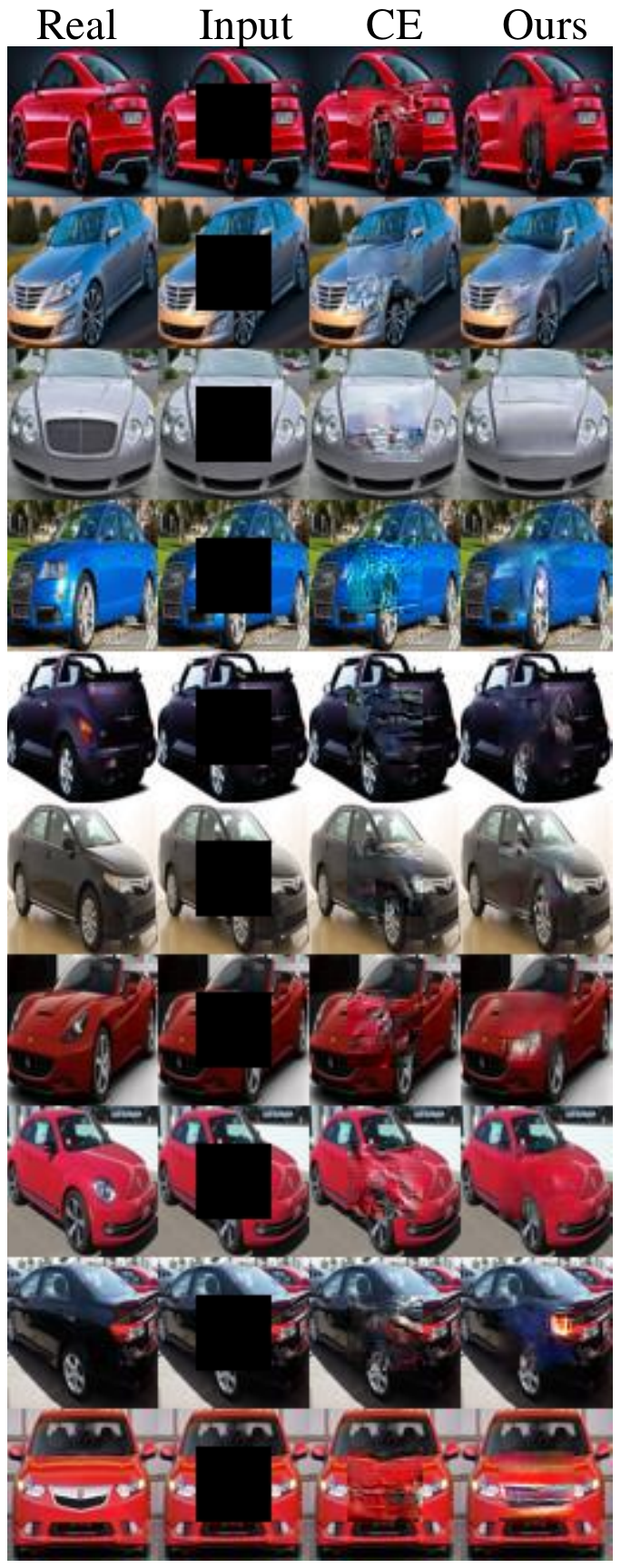} \hspace{0.25cm}
		\includegraphics[scale=1.1]{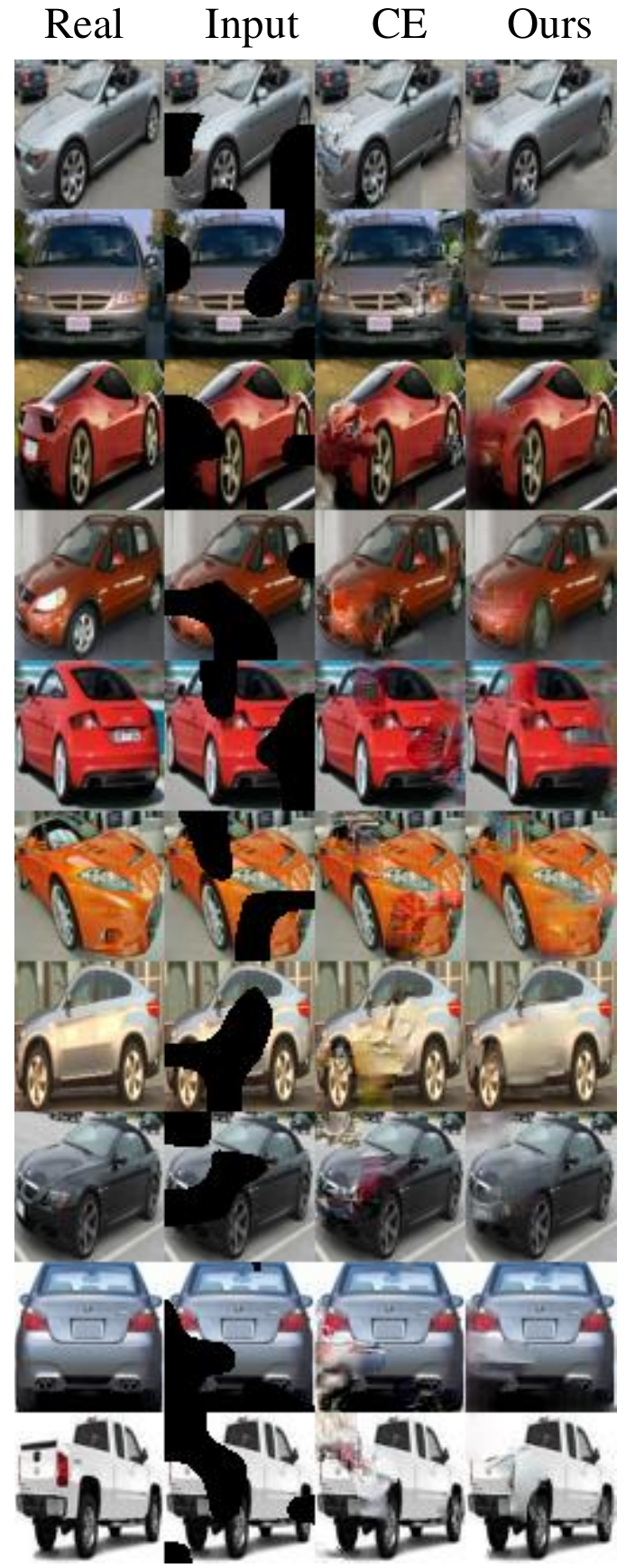}
		\caption{Additional results on the car dataset.}
\end{figure}

\begin{figure}[t]
\centering
		\includegraphics[scale=1.1]{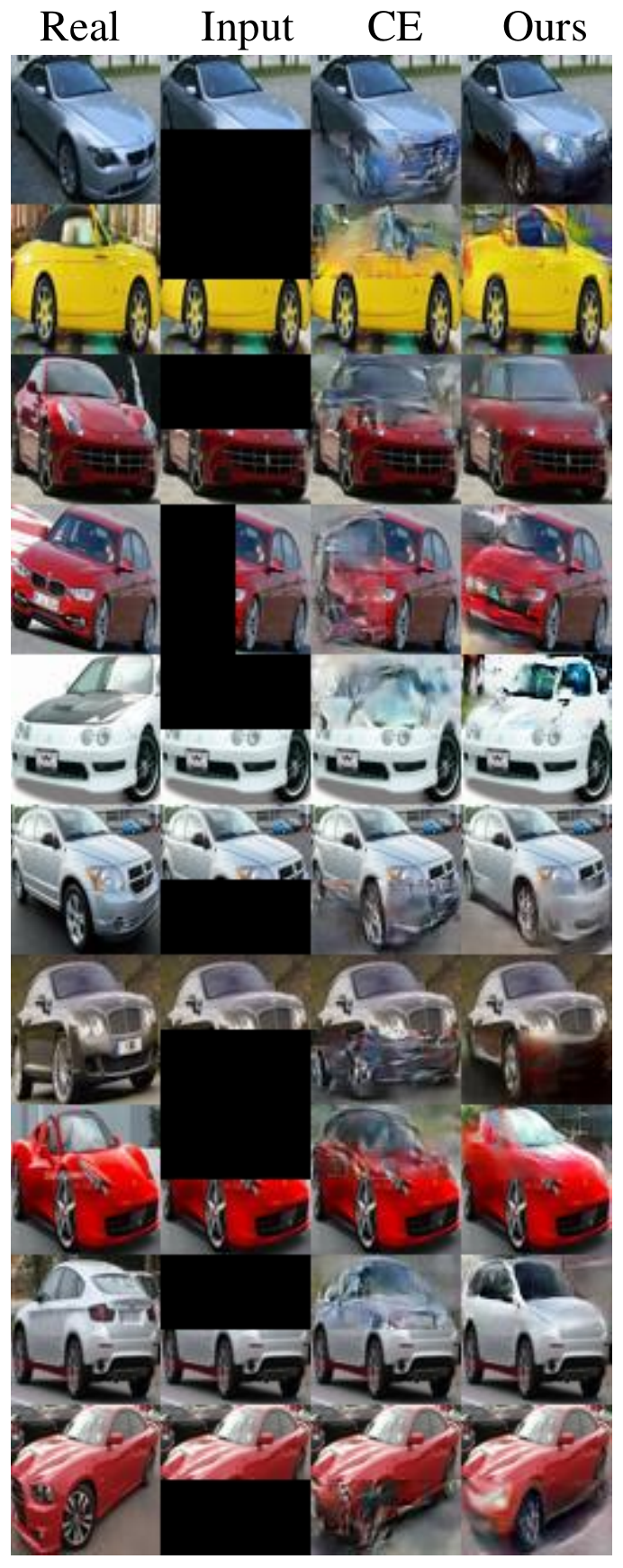} \hspace{0.25cm}
		\includegraphics[scale=1.1]{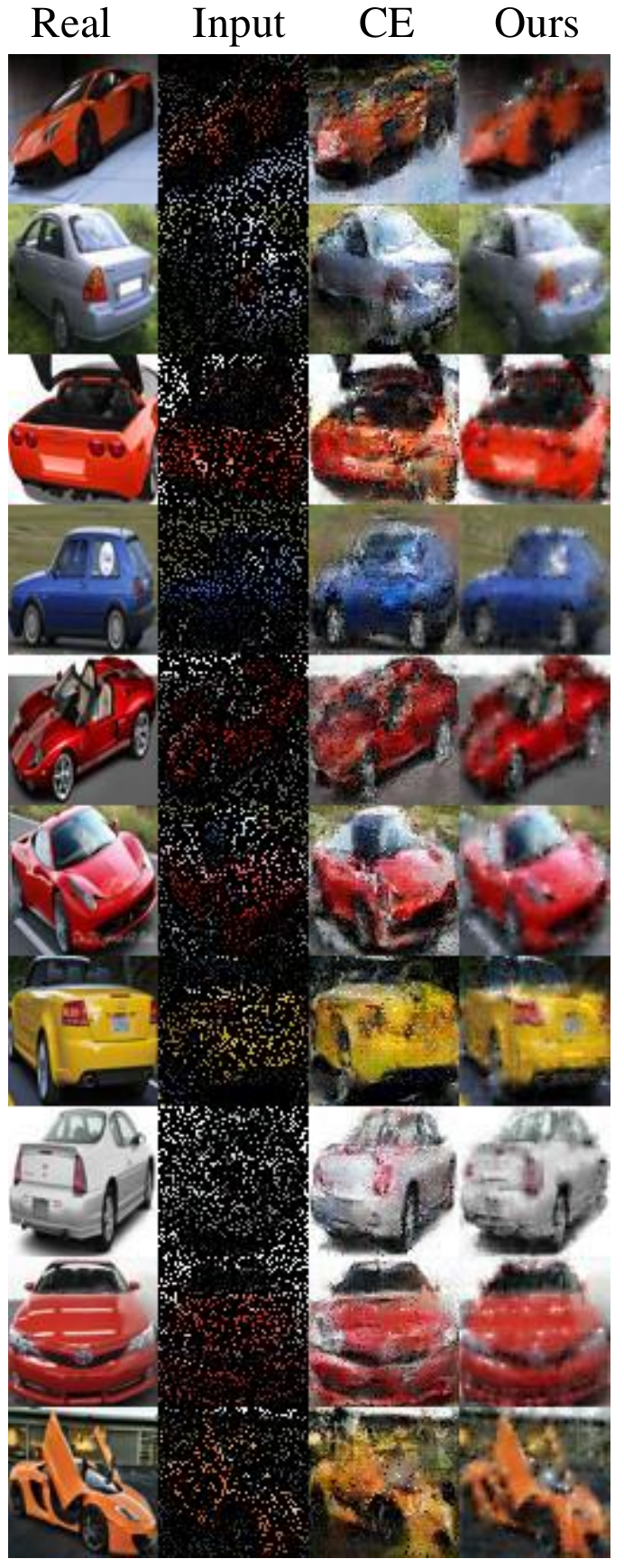}
		\caption{Additional results on the car dataset.}
\end{figure}

\end{document}